\documentclass[11pt,a4paper]{article}

\usepackage[style=ieee, backend=biber]{biblatex}
\usepackage{lscape}
\usepackage{colortbl}
\usepackage{booktabs}
\usepackage[utf8]{inputenc}
\usepackage[T1]{fontenc}
\usepackage{authblk}
\usepackage{titlesec}
\titleformat{\paragraph}
{\normalfont\normalsize\bfseries}{\theparagraph}{1em}{}
\titlespacing*{\paragraph}
{0pt}{3.25ex plus 1ex minus .2ex}{1.5ex plus .2ex}
\usepackage{mathrsfs}
\usepackage[dvips]{epsfig}
\usepackage{graphicx}
\usepackage{amssymb, graphics, amsmath, amsthm, amsopn, amstext, amsfonts, algorithm, algorithmic}
\usepackage{subcaption}
\usepackage{fancyhdr}
\usepackage{lipsum} 
\usepackage{booktabs} 
\usepackage{tabularx}
\usepackage{caption}
\usepackage{adjustbox} 
\usepackage{rotating}
\usepackage{color}
\usepackage[left=3cm,right=2cm,top=3cm,bottom=2cm]{geometry}

\usepackage{multirow}
\usepackage{multicol}
\usepackage{cases}
\usepackage{makecell}   
\usepackage{placeins}

\newcommand{\Argmin}{\mbox{Argmin}}

\newcommand{\Jcal}{\mathcal J}

\newcommand{\diag}{{\rm {\mbox{\bf{diag}}}}}

\newcommand{\mycut}[1]{{}}

\newcommand{\ve}{\mbox{vec}}
\newcommand{\hv}{\mbox{hvec}}

\newcommand{\RNum}[1]{\uppercase\expandafter{\romannumeral #1\relax}}
\newtheorem{corollary}{Corollary}[section] 
\newtheorem{definition}{Definition}[section]

\makeatletter
\def\thanks#1{\protected@xdef\@thanks{\@thanks
        \protect\footnotetext{#1}}}
\makeatother
\newcommand{\keywords}[1]{\textbf{\textit{keywords}}: #1}

\begin{document}

\title{\textbf{
$\ell_0$-Regularized Quadratic Surface Support Vector Machines
}}

\author[1]{Ahmad Mousavi$^*$\thanks{Emails: mousavi@american.edu, zandv003@umn.edu, and zgao7@charlotte.edu.
}} 
\affil{Department of Mathematics and Statistics, American University, Washington, DC, USA}
\author[2]{Ramin Zandvakili\thanks{}}
\affil[2]{Carlson School of Management, University of Minnesota, Minneapolis, MN, USA}
\author[3]{Zheming Gao\thanks{}}
\affil[3]{Department of Industrial and Systems Engineering, University of North Carolina at Charlotte, Charlotte, NC, USA}

\date{}
\maketitle
\begin{abstract}
Kernel-free quadratic surface support vector machines (QSVM) have recently gained traction due to their flexibility in modeling nonlinear decision boundaries without relying on kernel functions. However, the introduction of a full quadratic classifier significantly increases the number of model parameters, scaling quadratically with data dimensionality, which often leads to overfitting and makes interpretation difficult.
To address these challenges, we propose sparse variants of the QSVM by enforcing a cardinality constraint on the model parameters. While enhancing generalization and promoting sparsity, leveraging the $\ell_0$-norm inevitably incurs additional computational complexity. To tackle this, we develop a penalty decomposition algorithm capable of producing solutions that provably satisfy the first-order Lu-Zhang optimality conditions. 
We show that the subproblems arising within the algorithm either admit closed-form solutions or can be solved efficiently through dual formulations, which contributes to the method’s overall effectiveness. Besides, we analyze the convergence behavior of the algorithm under both loss settings. In addition, the numerical experiments on public benchmark datasets indicate that the proposed model is competitive with commonly used SVM variants and produces sparse solutions as expected. Moreover, its strong performance on real-world credit datasets demonstrates its potential for credit scoring applications.
\end{abstract}


\section{Introduction} \label{sec: intro}
Soft-margin support vector machines (SVMs) remain a fundamental tool for binary classification, effectively balancing margin maximization with error tolerance~\cite{lyaqini2024primal, ke2023maximal, chen2021equivalence}. In parallel, the development of sparse linear models has addressed the growing need for interpretability and overfitting control by deliberately restricting model complexity~\cite{chen2022jointly, landeros2023algorithms, xie2014sparse, pappu2015sparse, moosaei2023sparse1}. While valuable, the utility of linear sparsity is inherently bounded by its inability to model nonlinear relationships. Kernel methods were introduced to SVMs to resolve this, enabling the discovery of complex patterns in a transformed feature space. This advance, however, comes at a price. The resultant models frequently lose the transparency of their linear counterparts, becoming difficult to interpret. Furthermore, their effectiveness is heavily contingent on the careful, and often non-trivial, selection of the kernel and its associated hyperparameters, a process that can be computationally expensive and hinder practical applicability.

Kernel-free quadratic surface support vector machines (QSVMs) were developed to preserve model interpretability while avoiding the complexities of kernelization. These methods learn quadratic decision boundaries directly, optimizing an approximate geometric margin and maintaining robustness to noise and outliers~\cite{lu2013sparse}. Their explicit functional form offers additional flexibility, such as the ability to incorporate prior knowledge in the form of Universum data. Variants like twin QSVMs further enhance scalability by decomposing the primary problem into smaller subproblems~\cite{moosaei2023sparse}. Despite these advantages, a fundamental limitation persists: the number of parameters in a quadratic model grows quadratically with the input dimension, i.e., $\mathcal{O}(n^2)$. This quadratic scaling renders such models prone to overfitting and restricts their practicality, even on moderately sized datasets which is a challenge that subsequent research has sought to address.

The existing work further restricts the quadratic weight matrix to be diagonal~\cite{gao2022kernel}, which reduces the number of trainable parameters to $\mathcal{O}(n)$ and alleviates the overfitting problem. However, this simplification rests on a strong assumption that feature interactions are negligible. In practice, this assumption is seldom valid, and discarding pairwise correlations can lead to the loss of informative structure, resulting in underfitting. To navigate this trade-off between model complexity and representational power, various regularization techniques have been explored. $\ell_1$ norm regularization promotes sparsity and enhances interpretability, yet it may yield nonunique solutions~\cite{mousavi2022quadratic, moosaei2023sparse, xiao2023kernel,si2025kernel,zhou2023kernel}. $\ell_2$ norm regularization effectively curbs overfitting but does not inherently encourage sparsity~\cite{wang20252}. More flexible approaches, such as $\ell_p$ norm regularization, allow finer control over the degree of sparsity, though this comes at the cost of careful tuning of the parameter $p$~\cite{yang2025kernel}.

Sparse recovery theory provides a nuanced understanding of how different choices of $p$ in $\ell_p$ regularization affect the structure of the solution. Values of $p>1$ tend to yield dense solutions, offering little in the way of sparsity~\cite{shen2018least}. The case $p=1$ preserves convexity and promotes sparsity, yet it remains less aggressive in this regard than norms with $0<p<1$. Within the nonconvex regime, values of $p$ in the range $[1/2, 1)$ often produce the sparsest solutions overall, with diminishing returns as $p$ decreases further~\cite{zong2012representative}. Against this backdrop, $\ell_0$ norm regularization occupies a unique position. Unlike any $\ell_p$ surrogate with $p>0$, it enables exact control over the number of nonzero parameters in the solution. This capability delivers not only precise sparsity but also stronger theoretical guarantees for recovery. Recent algorithmic advances have rendered direct $\ell_0$ optimization increasingly practical. Methods such as penalty decomposition~\cite{lu2022penalty}, greedy algorithms~\cite{gupta2024greedy}, and branch and bound techniques~\cite{hazimeh2022sparse} demonstrate that convex relaxations are no longer necessary and are often suboptimal for $\ell_0$ regularized problems.

Building on these insights, we develop novel kernel-free QSVM models that employ $\ell_0$ regularization to achieve exact sparsity in both the quadratic and linear coefficients. Since the proposed $\ell_0$ regularized models are intractable, an efficient algorithm is designed for their implementation. Numerical experiments are conducted to validate the sparsity and the classification capability of the proposed models. The contributions of this research are summarized as follows.
\begin{itemize}
    \item Our method combines the representational capacity of quadratic decision boundaries with the simplicity of sparse models, offering both powerful nonlinear classification and clear interpretability. Different from the sparsity yielded by $\ell_1$ regularizations \cite{huang2015two, mousavi2022quadratic}, the $\ell_0$ constraint provides solutions with at most $k$ non-zero elements, giving direct control over model complexity while automatically performing feature selection.

    \item To address the computational challenges brought up by the $\ell_0$ regularization, we develop a penalty decomposition algorithm that recasts the original problem into a sequence of more tractable subproblems. By introducing an auxiliary variable that decouples the $\ell_0$-norm constraint, the subproblem alternates between analytically tractable updates and efficiently solvable ones via duality theory. In addition, the convergence of the proposed penalty decomposition algorithm is rigorously discussed.

    \item Extensive numerical experiments are conducted to evaluate the effectiveness of the proposed models in classification tasks. The sparsity characteristics of the models and the influence of key parameters are systematically examined through computational analysis. In addition, the proposed model with quadratic loss is applied to real-world credit scoring datasets, demonstrating strong predictive performance and enhanced interpretability in practical credit risk assessment applications.
\end{itemize}

The remainder of the paper is structured as follows. Section~\ref{sec: related_work} reviews relevant literature. Section~\ref{sec: model_and_methodology} presents our $\ell_0$-regularized QSVM models and describes the decomposition algorithm. Section~\ref{sec: numerical_experiments} presents the numerical experiment results as well as the application of the proposed models to credit scoring. Section~\ref{sec: conclusion} concludes with a discussion of future directions.

\section{Preliminaries and Related Work} \label{sec: related_work}
This section briefly reviews several quadratic surface support‑vector machine (QSVM) models.  
Consider a binary classification task with a training set
\begin{equation*}
 \mathcal D = \{\left({x_i} , y_i\right)_{i=1,\dots,m} \ \left| \ {x_i}  \, \in \, \mathbb{R}^n, \ y_i \in \{-1, 1\}\right.
\},
\end{equation*}
where $m$ is the sample size, $n$ is the number of features, and $y_i$ is the label of $x_i$. 
$\mathcal D$ is called quadratically separable \cite{lu2013sparse} if there exist $W\in \mathbb{R}^{n\times n},    b \in \mathbb{R}^n,\, \text{and} \, c \in \mathbb{R}$ if $y_i(x_i^{\top} W {x_i}+b^{\top} {x_i}+c)>0$ for every $i\in [m]$.
A linearly separable dataset is a special case of quadratic separability obtained by setting \(W=0\).

Standard SVMs seek a hyperplane that approximately separates a training set. In many real‑world binary classification problems, however, data exhibit non‑linear patterns, making linear separation difficult. Kernel methods overcome this limitation by mapping the data into a higher‑dimensional feature space in which non‑linear relationships appear linear. In that space, SVMs can identify optimal hyperplanes that more effectively capture complex structures. Consequently, kernel techniques markedly improve SVM performance on non‑linear classification tasks and are widely applied in machine learning, pattern recognition, and data mining \cite{chandra2021survey}.

Despite their advantages, kernel methods also have notable limitations. They are computationally expensive because pairwise similarity calculations scale poorly with large datasets. Moreover, as the dimensionality of the feature space grows, kernel models become prone to overfitting. Choosing a kernel function demands domain expertise or extensive experimentation, and the resulting classifiers are less interpretable than their linear counterparts. Sensitivity to hyper‑parameter tuning and limited scalability further underscore the trade‑offs associated with kernel methods in complex classification tasks \cite{hofmann2008kernel}.

Therefore, a recent method, introduced by Dagher \cite{dagher2008quadratic} and further refined by Luo et al. \cite{lu2013sparse}, seeks to explicitly separate data using a quadratic classifier directly in the original input space, avoiding any kernel‑induced feature mapping: \[ f_{W,b,c}(x) = \frac{1}{2} x^{\top} W x + b^{\top} x + c = 0.
\]
The soft‑margin variant of this model, which penalizes misclassifications to accommodate noise and outliers, is formulated as follows:
\begin{equation}\tag{QSVM}\label{QSVM}
\begin{aligned}
\min_{W, b, c,   \xi} \quad & \sum_{i=1}^m \| W {x_i}  +   b\|_2^2 + {\mathcal{C}} \sum_{i=1}^m    \xi_i \\
\mbox{s.t.} \quad & y_i( \frac{1}{2} x_i^{\top} W {x_i}  + b^{\top}x_i + c ) \ge 1-    \xi_i, \quad i=1,\dots,m, \\
           & W\in \mathbb S_n, \,   b \in \mathbb{R}^n, \, c \in \mathbb R, \,   \xi \in \mathbb{R}^m_+.
\end{aligned}
\end{equation}
Notice that adopting a kernel‑free quadratic surface framework dramatically enlarges the model’s degrees of freedom: the number of parameters grows on the order of $\mathcal{O}(n^{2})$, compared with the $\mathcal{O}(n)$ parameters in a standard linear SVM. While this added flexibility can improve the fit to the training data, it also raises the risk of overfitting and may diminish generalization to unseen samples. The danger is especially acute for linearly separable datasets, where one would prefer the QSVM to revert to a linear separator—yet QSVM offers no such guarantee. A widely used remedy is sparsity‑promoting $\ell_{1}$‑norm regularization; for example, Mousavi et al. \cite{mousavi2022quadratic} embed an $\ell_{1}$ penalty directly into the QSVM objective:
\begin{equation} \tag{$\ell_1$-QSVM}\label{L1-QSVM}
\begin{aligned}
\min_{W, b, c,   \xi} \quad & \sum_{i=1}^m \| W {x_i}  +   b\|_2^2 + {\mathcal{C}}_1 \sum_{1\le i\le j\le n} | W_{ij}| + {\mathcal{C}}_2 \sum_{i=1}^m    \xi_i \\
\mbox{s.t.} \quad & y_i( \frac{1}{2} x_i^{\top} W {x_i}  + b^{\top}x_i + c ) \ge 1-    \xi_i; \quad \forall i\in [m], \\
           & W\in \mathbb S_n, \,   b \in \mathbb{R}^n, \,  c \in \mathbb R, \,   \xi \in \mathbb{R}^m_+,
\end{aligned}
\end{equation}
where ${\mathcal{C}}>0$ is a positive penalty for incorporating noise and outliers. 
The authors demonstrate that the above model reduces to the standard SVM when $\lambda$ is sufficiently large. Other $\ell_0$-norm surrogates, such as $\ell_p$ norms, for $p \in (0,1)$, have also been explored in the literature on linear SVMs, feature selection, and $K$-support vector classification regression. However, the non-convexity of these models poses significant challenges for algorithm design \cite{yao2017sparse, blanco2020lp,ma2019sparse}.

Although $\ell_{p}$ norms with $0<p\le 1$ promote sparsity, only the $\ell_{0}$ norm enforces it directly by minimizing the number of non‑zero parameters. In compressive sensing, for example, $\ell_{0}$ achieves higher compression rates than either the convex ($p=1$) or non‑convex ($0<p<1$) $\ell_{p}$‑norm relaxations, making it a superior tool for sparse representation. Motivated by this property, we next study sparse $\ell_{0}$‑norm, kernel‑free quadratic surface SVMs and present a penalty decomposition algorithm that solves these models efficiently.

Kernel-free quadratic surface models differ from conventional quadratic program formulations in quadratic optimization. Hence, several notations and definitions must be introduced to rectify this and align them accordingly. We start with a square matrix $A = [a_{ij}]_{i=1,\dots,n;j=1,\dots,n}\in \mathbb R^{n\times n}$. Its vectorization is straightforwardly represented as $\ve(A) := \left[a_{11},\dots, a_{n1},a_{12},\dots,a_{n2},\dots,a_{1n},\dots,a_{nn}\right]^{\top} \in \mathbb{R}^{n^2}$. However, if $A$ is symmetric, $\ve(A)$ carries redundant data, hence we opt for its half-vectorization:
\[
\hv(A) := \left[a_{11},\dots, a_{n1},a_{22},\dots,a_{2n}, \dots,a_{nn}\right]^{\top} \in \mathbb{R}^{\frac{n(n+1)}{2}}.
\]
For any $n\in \mathbb{N}$, there exists a unique elimination matrix $L_n \in \mathbb{R}^{\frac{n(n+1)}{2} \times n^2}$ such that $L_n \ve(A) = \hv(A)$ for any $A \in S_n$, where $S_n$ denotes the set of symmetric matrices. Additionally, this elimination matrix $L_n$ has a full row rank. Conversely, for any $n\in \mathbb{N}$, there exists a unique duplication matrix $D_n \in \mathbb{R}^{n^2 \times \frac{n(n+1)}{2}}$ such that $D_n \hv(A) = \ve(A)$ for any $A \in S_n$, and $L_nD_n=I_{\frac{n(n+1)}{2}}$.

\begin{definition} \label{def: definitios}
For $i\in [m],$ let
\begin{eqnarray}
 s_i := &  \frac{1}{2}\hv( x_i x_i^{\top}), \quad
r_i :=  [s_i;x_i], \quad
w := \hv(W), \quad
z: = [w;b],\quad
V:= \begin{bmatrix} I_{\frac{n(n+1)}{2}} &   0_{\frac{n(n+1)}{2}\times n} \end{bmatrix}, \nonumber \\
X_i := &  I_n \otimes x_i^{\top}, \quad 
M_i :=   X_i D_n, \quad 
H_i:= 
\begin{bmatrix} M_i &   I_{n}\end{bmatrix}, \quad G :=  2 \sum_{i=1}^{m} H_i^{\top} H_i, \quad
X:= [x_1, x_2,\dots,x_m]^{\top}. \nonumber
\end{eqnarray}
\end{definition}
Hence, simply $G\succeq 0$. And, for fixed $i\in[m]$, we have the following equations:
\begin{equation*}
\left\{\begin{array}{ll}
Wx_i  =  X_i\ve( W) \, = X_i   D_n \hv(  W) \, = \, M_i \hv(  W) \, = \, M_iw, & \\
W x_i +    b  =  \, M_iw +  I_nb \, = \, H_iz, & \\
\frac{1}{2}x_i^{\top}W x_i + b^{\top}x_i + c=  z^{\top} r_i+ c. 
\end{array} \right.
\end{equation*}
Thus, $$ \sum_{i=1}^m \|Wx_i+b\|_2^2 =\sum_{i=1}^m \left(H_i    z\right)^{\top}\left(H_i z\right) \,= \,    z^{\top} \big(\sum_{i=1}^{m} H_i^{\top} H_i\big)z=\frac{1}{2}z^{\top}Gz.$$

\section{$\ell_0$-Regularized Quadratic Surface SVMs and A Penalty Decompotion Algorithm} \label{sec: model_and_methodology}

Empirical evidence suggests that many real-world datasets, being governed by underlying physical laws, are effectively controlled by only a few dominant factors. This makes $\ell_0$-regularized models particularly appealing for capturing their essential structure. By isolating these critical components, $\ell_{0}$ regularization delivers greater interpretability and practical utility. Conversely, $\ell_{p}$ surrogates with $0<p\le 1$ merely approximate sparsity: they shrink coefficients toward—but rarely all the way to—zero, and the degree of sparsity is controlled only indirectly through a tuning parameter. Because this parameter cannot set the exact number of non‑zero entries, $\ell_{p}$ regularization often fails to capture the true zero pattern. These shortcomings highlight the benefits of employing $\ell_{0}$ regularization when precise sparsity is crucial. However, the inclusion of a sparsity constraint renders the model intractable, necessitating the development of an efficient solution method. To address this, we propose a penalty decomposition algorithm and demonstrate how its subproblems can be solved effectively. We will also study the convergence of this algorithm.

To mitigate overfitting arising from the over‑parameterization of the symmetric matrix $W$ and vector $b$ in the quadratic classifier  $f$, we impose sparsity constraints on these parameters. This restriction lowers model complexity and improves generalization. Accordingly, we examine two sparse 
$\ell_0$-norm kernel‑free quadratic surface SVM formulations. The first employs the hinge loss and is expressed as follows:

\begin{equation} \tag{$\ell_0$-QSVM} \label{L0-QSVM}
\begin{aligned}
\min_{W, b, c,   \xi} \quad &    \sum_{i=1}^m \| W {x_i}+b\|_2^2+{\mathcal{C}} \sum_{i=1}^m    \xi_i \\
\mbox{s.t.} \quad & y_i( \frac{1}{2} x_i^{\top} W {x_i}  + b^{\top}x_i + c ) \ge 1-    \xi_i; \quad \forall i \in [m],\\
           &\|[\hv(W);b]\|_0 \le k,\\
           & W\in \mathbb S_n, \,   b \in \mathbb{R}^n, \,  c \in \mathbb{R}, \,   \xi \in \mathbb{R}^m_+.
\end{aligned}
\end{equation}
Using the quadratic loss function, we derive the least-squares version of the model as follows:
\begin{equation} \tag{LS-$\ell_0$-QSVM} \label{LS-L0-QSVM}
\begin{aligned}
\min_{W, b, c,   \xi} \quad & \sum_{i=1}^m\|Wx_i+b\|_2^2+{\mathcal{C}} \sum_{i=1}^m    \xi^2_i \\
\mbox{s.t.} \quad & y_i( \frac{1}{2} x_i^{\top} W {x_i}  + b^{\top}x_i + c ) = 1-    \xi_i; \quad \forall i\in [m], \\
           & \|[\hv(W);b]\|_0 \le k,\\
           & W\in \mathbb S_n, \,   b \in \mathbb{R}^n, \,  c \in \mathbb{R}, \,   \xi \in \mathbb{R}^m,
\end{aligned}
\end{equation}
To present a penalty decomposition algorithm that works for both proposed models, with different corresponding subproblems, we introduced the following unified model:
\begin{equation*} %
\begin{aligned}
\min_{W\in \mathbb S_n, \,   b \in \mathbb{R}^n, \,  c \in \mathbb{R}} \quad &   
\sum_{i=1}^m \|Wx_i+b\|_2^2
+{\mathcal{C}} \sum_{i=1}^m    \mathrm H(1-y_if_{W,b,c}(x_i))  \qquad 
\mbox{s.t.} \qquad
\|[\hv(W);b]\|_0 \le k,
\end{aligned}
\end{equation*}
which reduces to \ref{L0-QSVM} for the well-received hinge loss $\mathrm H(t)= \max(t,0)$ and reduces to \ref{LS-L0-QSVM'} for the quadratic loss  $\mathrm H(t) = t^2$. Based on the notation introduced in the previous section, the above model is equivalent to:
\begin{equation} \label{generic-L0-quadratic}
\begin{aligned}
\min_{z\in \mathbb R^{\frac{n(n+1)}{2}+n},  c \in \mathbb{R}} \quad &    
\frac{1}{2}z^{\top}Gz
+{\mathcal{C}} \sum_{i=1}^m    \mathrm H(1-y_i(z^{\top}r_i+c))  
\qquad \mbox{s.t.} \qquad
           \|z\|_0 \le k.
\end{aligned}
\end{equation}
This unified model is still intractable, and therefore, we next apply an effective penalty decomposition method to solve it. After introducing a new variable $u$, this problem becomes:
\begin{equation*} 
\begin{aligned}
\min_{z,u\in \mathbb R^{\frac{n(n+1)}{2}+n},  c \in \mathbb{R}} \quad &    
\frac{1}{2}z^{\top}Gz
+{\mathcal{C}} \sum_{i=1}^m    \mathrm H(1-y_i(z^{\top}r_i+c))  
\qquad \mbox{s.t.} \qquad
           \|z\|_0 \le k \qquad \mbox{and} \qquad z-u=0.
\end{aligned}
\end{equation*}
Thus, by penalizing the last constraint above, the problem \ref{generic-L0-quadratic} can be addressed by solving a sequence of penalty subproblems as follows:
\begin{equation}
\tag{$P_\rho$} \label{penalty_subproblem_generic}
\begin{aligned}
\min_{z,u\in \mathbb R^{\frac{n(n+1)}{2}+n},  c \in \mathbb{R}} \quad &    
q_\rho(z,c,u):=
\frac{1}{2}z^{\top}Gz
+{\mathcal{C}} \sum_{i=1}^m  \mathrm H(1-y_i(z^{\top}r_i+c))+\frac{1}{2}\rho\|z-u\|_2^2 
\qquad \mbox{s.t.} \qquad 
           \|u\|_0 \le k.
\end{aligned}
\end{equation}
Suppose we have a feasible point denoted as $(z,c)^{\text{feas}}$ of \ref{generic-L0-quadratic} in hand. Then, let:
\begin{equation*} \label{def: upsilon}
\Upsilon\ge\max\{f((z,c)^{\text{feas}}), \min_{z,c} q_{\rho^{(0)}}(z,c,u^{(0)}_0)\}> 0.
\end{equation*}
We solve the above penalty subproblem for a fixed positive $\rho$ via a block coordinate descent method. 
The stopping criterion for the inner loop is the following:
\begin{equation} \label{BCD-practical-stopping-criterion}
\max
\left\{
\frac{\|z_l-z_{l-1}\|_\infty}{\max \left(\|z_l\|_\infty,1 \right)},
\frac{\|c_l-c_{l-1}\|_\infty}{\max \left(\|c_l\|_\infty,1 \right)},\frac{\|u_l-u_{l-1}\|_\infty}{\max \left(\|u_l\|_\infty,1 \right)},
\right\}
\le
\epsilon_I.
\end{equation}
And for the outer loop is:
\begin{equation} \label{outer loop stopping criteria}
\|z^{(j)}-u^{(j)}\|_{\infty}
\le
\epsilon_O.
\end{equation}
Therefore, we present the following penalty decomposition algorithm for solving (\ref{generic-L0-quadratic}).

\begin{algorithm}[H]
\caption{$\ell_0$-Regularized QSVM Penalty Decomposition}
\begin{algorithmic}[1]
\label{algo: L0-penalty}
\STATE Inputs:  $ \rho^{(0)}>0, \beta>1, k$, and  $u^{(0)}_0$ such that $\|u^{(0)}_0\|_0 \le k$.
\STATE $j\gets 0$.
\REPEAT
 \STATE $l\gets 0$. 
\REPEAT 
\STATE 
\vspace{.08cm}
$(z^{(j)}_{l+1},c^{(j)}_{l+1}) \gets \Argmin_{(z, c)\in \mathbb R^{\frac{n(n+1)}{2}+n}\times \mathbb R} \ q_{\rho^{(j)}}(z, c,u^{(j)}_l).$  
\STATE 
\vspace{.08cm}
$u^{(j)}_{l+1} \gets \Argmin_{\{u|\|u\|_0 \le k\}} \ q_{\rho^{(j)}}(z^{(j)}_{l+1},c^{(j)}_{l+1},u).$ 
\vspace{.08cm}
 \STATE $l \gets l+1$.
 \UNTIL{stopping criterion (\ref{BCD-practical-stopping-criterion}) is met.} 
 \STATE \vspace{.08cm}
$(z^{(j)},c^{(j)},u^{(j)})\gets
(z^{(j)}_l,c^{(j)}_l,u^{(j)}_l)$.
\vspace{.08cm}
\STATE   $\rho^{(j+1)} \gets \beta\cdot \rho^{(j)}$.
\STATE 
If $\min_{(z, c)\in \mathbb R^{\frac{n(n+1)}{2}+n}\times \mathbb R}\ q_{\rho^{(j+1)}}(z,c,u^{(j)})> \Upsilon$, then $u_0^{(j+1)}\gets z^{\text{feas}}$.
Otherwise,
$u^{(j+1)}_0 \gets u^{(j)}_{l}$.
\STATE  $j \gets j+1$.
\UNTIL{stopping criterion (\ref{outer loop stopping criteria}) is met}. 
\end{algorithmic}
\end{algorithm}

We now turn our attention to solving the restricted subproblems within the algorithm as efficiently as possible.
Observe that the associated subproblem for $u$  is:
\begin{equation} 
\label{pr: pu} 
\min_{u\in \mathbb{R}^{\frac{n(n+1)}{2}+n}} \ \|z-u\|_2^2 \qquad \mbox{s.t.}  \qquad   \|u\|_0 \le k;
\end{equation}
which obtains a solution
\begin{equation} \label{sol-pu}
u=[z_\Jcal;0],
\end{equation}
where $\Jcal$ contains indices of $k$ largest components of $z$ in absolute value,  that is, Step 7 of Algorithm \ref{algo: L0-penalty} has a closed-form solution.
Next, we start discussing how to tackle the associated subproblem to $z$ for each of the loss functions.

\subsection{$\ell_0$-Regularized Quadratic Surface SVM Model}
For the hinge loss function $\mathrm H(t)=\max(0,t)$, the sparse quadratic model \ref{generic-L0-quadratic} reduces to \ref{L0-QSVM}, which after using the notation in Section \ref{sec: related_work} is equivalent to:
\begin{equation} \tag{$\ell_0$-QSVM'} \label{L0-QSVM'}
       \begin{aligned}
          \min_{z,c,\xi} \quad &  \frac{1}{2} { z}^{\top}Gz+{\mathcal{C}} \sum_{i=1}^m \xi_i\\
          \mbox{s.t.} \quad & y_i(z^{\top} r_i + c ) \ge 1-\xi_i; \quad \forall i\in [m], \\
           &   \|z\|_0 \le k.
          \\
           &   z\in \mathbb{R}^{\frac{n(n+1)}{2}+n},c\in \mathbb{R}, \xi \in \mathbb R^m_+.
      \end{aligned}
\end{equation}
Recalling \ref{penalty_subproblem_generic}, the subproblem associated to $z$ for the hinge loss function becomes:
\begin{equation*}
\begin{aligned}
\min_{z\in \mathbb R^{\frac{n(n+1)}{2}+n},  c \in \mathbb{R}} \quad &    
\frac{1}{2}z^{\top}Gz
+{\mathcal{C}} \sum_{i=1}^m    \max(1-y_i(z^{\top}r_i+c),0)+\frac{1}{2}\rho\|z-u\|_2^2,
\end{aligned}
\end{equation*} 
which is equivalent to the following feasible and bounded below convex quadratic optimization problem: 
\begin{equation} \tag{$P^h_{\rho,(z,c,\xi)}$}
\label{Pz-hinge}
\begin{aligned}
          \min_{z,c,\xi} \quad &  \frac{1}{2} {z}^{\top}(G+\rho I) z-\rho u^{\top}z+{\mathcal{C}} \sum_{i=1}^m \xi_i
          \\
          \mbox{s.t.} \quad & y_i(z^{\top} r_i + c ) \ge  1-\xi_i;\quad \forall i\in [m] \\
           &   z\in \mathbb{R}^{\frac{n(n+1)}{2}+n},c\in \mathbb{R}, \xi \in \mathbb R^m_+, .
\end{aligned}
\end{equation}
Even though a standard solver can be used for this problem, standard practice in the SVM literature encourages applying the strong duality theorem in convex programming. Thus, let us bring the Lagrangian function of this model:
\begin{eqnarray}
L(z,c,\xi,\alpha) &=&  
\frac{1}{2} {z}^{\top}(G+\rho I)z-\rho u^{\top}z+{\mathcal{C}} \sum_{i=1}^m \xi_i
-\sum_{i=1}^m\alpha_i(y_i(z^{\top}r_i+c)-1+\xi_i)\nonumber \\
&=&
\frac{1}{2} {z}^{\top}(G+\rho I)z -z^{\top}\Big(\sum_{i=1}^m\alpha_iy_ir_i+\rho u\Big)+\sum_{i=1}^m({\mathcal{C}}-\alpha_i)\xi_i-c\sum_{i=1}^m \alpha_iy_i.
\nonumber
\end{eqnarray}
Therefore, the dual problem of \ref{Pz-hinge} becomes:
\begin{equation} \tag{$D_{\alpha}$}
\label{dual-Pz-hinge}
\begin{aligned}
          \min_{\alpha\in \mathbb R^m} \quad &  -\sum_{i=1}^m\alpha_i+
          \frac{1}{2}\Big(\sum_{i=1}^my_i\alpha_ir_i+\rho u\Big)^{\top}(G+\rho I)^{-1}\Big(\sum_{j=1}^my_j\alpha_jr_j+\rho u\Big)
          \\
          \mbox{s.t.} \quad & \sum_{i=1}^m y_i\alpha_i=0 \qquad \mbox{and} \qquad 0\le \alpha \le \textbf{1}{\mathcal{C}}.\\
\end{aligned}
\end{equation}
Therefore, the Karush-Kuhn-Tucker (KKT) conditions are:
\begin{equation*}
\left\{\begin{array}{ll}
z  =   (G+\rho I)^{-1} \big(\sum_{i=1}^m\alpha_iy_ir_i+\rho u\big) & \\
\alpha_i(1-\xi_i-y_i(z^{\top} r_i + c ))=0; \quad \forall i \in [m] & \\
({\mathcal{C}}-\alpha_i)\xi_i=0; \quad \forall i \in [m] &\\
y_i(z^{\top} r_i + c ) \ge  1-\xi_i;  \quad \forall i \in [m] & \\
\sum_{i=1}^m y_i\alpha_i=0; \quad \forall i \in [m] & \\
\xi\in \mathbb R^m_+ \qquad \mbox{and} \qquad 0\le \alpha \le \textbf{1}{\mathcal{C}}, & \\
\end{array} \right.
\end{equation*}
which implies that, once a solution to the dual \ref{dual-Pz-hinge} is provided,  a solution to \ref{Pz-hinge} can be obtained from:
\begin{equation} \label{sol-Pz-hinge}
z  =   (G+\rho I)^{-1} \Big(\sum_{\alpha_i\ne 0}\alpha_iy_ir_i+\rho u\Big), \qquad 
c=\max_{y_i=1, \alpha_i>0} -z^{\top}r_i, \qquad
\xi_i =\left\{\begin{array}{ll}
 1-y_i(z^{\top}r_i+c) \quad \mbox{if} \quad \alpha_i>0, & \\
0 \quad \mbox{if} \quad  \alpha_i=0. & \\
\end{array} \right.
\end{equation}
\begin{corollary} \label{cor: 1}
For the hinge loss function, step 6 of Algorithm \ref{algo: L0-penalty} can be obtained by solving \ref{dual-Pz-hinge} and then applying \ref{sol-Pz-hinge}. 
\end{corollary}

\subsection{Least-squares $\ell_0$-Regularized Quadratic Surface SVM Model}
For the quadratic loss function $\mathrm H(t)=t^2$, the sparse quadratic model \ref{generic-L0-quadratic} reduces to the following least-squares $\ell_0$-regularized kernel-free quadratic surface SVM model \ref{LS-L0-QSVM}, which through the notations presented in Section \ref{sec: related_work} becomes:
\begin{equation} \tag{LS-$\ell_0$-QSVM'} \label{LS-L0-QSVM'}
       \begin{aligned}
          \min_{z,c,\xi} \quad &  \frac{1}{2} { z}^{\top}Gz+{\mathcal{C}} \sum_{i=1}^m \xi^2_i\\
          \mbox{s.t.} \quad & y_i(z^{\top} r_i + c ) = 1-\xi_i; \quad \forall i\in [m], \\
           &   \|z\|_0 \le k,
          \\
           &   z\in \mathbb{R}^{\frac{n(n+1)}{2}+n},c\in \mathbb{R}, \xi \in \mathbb R^m, .
      \end{aligned}
\end{equation}
The associated subproblem to $z$ in this case is equivalent to:
\begin{equation}\tag{$P_{(z,c)}^q$}
\label{Pz-hinge_quad}
\begin{aligned}
          \min_{z \in \mathbb{R}^{\frac{n(n+1)}{2}+n},c\in \mathbb{R}}
         &  \frac{1}{2} {z}^{\top}Gz+{\mathcal{C}} \sum_{i=1}^m (1-y_i(z^{\top}r_i+c))^2+\frac{1}{2}\rho \|z-u\|^2_2.
\end{aligned}
\end{equation}
By letting $A=[r_1, r_2, \dots, r_m]^{\top}$ and $D=\diag(y_1,y_2, \dots, y_m)$, the objective function of the latter is equal to $T=\frac{1}{2}z^{\top}(G+\rho I)z+{\mathcal{C}} \|\textbf{1}-D(Az+c\textbf{1})\|_2^2-\rho u^{\top}z+\rho\|u\|_2^2$. Hence,
\begin{eqnarray}
\frac{\partial T}{\partial z} &=&  
(G+\rho I)z-2{\mathcal{C}} A^{\top}D(\textbf{1}-D(Az+c\textbf{1}))-\rho u
\nonumber \\
\frac{\partial T}{\partial c} &=& -2{\mathcal{C}}\textbf{1}^{\top}D (\textbf{1}-D(Az+c\textbf{1}))
\nonumber
\end{eqnarray}
which can be integrated into the following:
\begin{equation*}
    \begin{bmatrix}
G+\rho I+2{\mathcal{C}} A^{\top}D^2A & 2{\mathcal{C}} A^{\top}D\textbf{1} \\
2{\mathcal{C}} \textbf{1}^{\top}D^2A & 2{\mathcal{C}} \textbf{1}^{\top}D^2\textbf{1} 
\end{bmatrix}
\begin{bmatrix}
z\\
c
\end{bmatrix}
=
\begin{bmatrix}
2{\mathcal{C}} A^{\top}D\textbf{1}+\rho u\\
2{\mathcal{C}} \textbf{1}^{\top}D\textbf{1},
\end{bmatrix}
\end{equation*}
which has a positive definite coefficient matrix such that:
\begin{equation} \label{sol-Pz-quad}
\begin{bmatrix}
z\\
c
\end{bmatrix}
=
    \begin{bmatrix}
G+\rho I+2{\mathcal{C}} A^{\top}D^2A & 2{\mathcal{C}} A^{\top}D\textbf{1} \\
2{\mathcal{C}} \textbf{1}^{\top}D^2A & 2{\mathcal{C}} \textbf{1}^{\top}D^2\textbf{1} 
\end{bmatrix}^{-1}
\begin{bmatrix}
2{\mathcal{C}} A^{\top}D\textbf{1}+\rho u\\
2{\mathcal{C}} \textbf{1}^{\top}D\textbf{1},
\end{bmatrix}
\end{equation}

\begin{corollary} \label{cor: 2}
For the quadratic loss function, step 6 of Algorithm \ref{algo: L0-penalty} is obtained from (\ref{sol-Pz-quad}). 
\end{corollary}
To conclude, we present two sparse $\ell_{0}$‑norm, kernel‑free quadratic surface SVMs that employ hinge and quadratic loss functions, respectively. Because the presence of a sparsity constraint renders these formulations NP‑hard, an efficient solution method is essential. We place both models within the unified framework~\eqref{generic-L0-quadratic} and develop a penalty decomposition algorithm that introduces an auxiliary variable $u$. The associated subproblem~\eqref{pr: pu} admits a simple closed‑form solution~\eqref{sol-pu}. For the hinge‑loss‑based model~\eqref{L0-QSVM}, the $z$-subproblem is solved most effectively via convex duality, with the primal variables recovered through~\eqref{sol-Pz-hinge}. In the quadratic‑loss‑based model~\eqref{LS-L0-QSVM}, the $z$-subproblem itself has the closed‑form solution given in~\eqref{sol-Pz-quad}. Consequently, each iteration of our algorithm remains computationally efficient while enforcing exact sparsity.

\subsection{Convergence of $\ell_0$-Regularized QSVM Penalty Decomposition Algorithm}

While convergence guarantees are well established for convex models, the presence of a nonconvex and nonsmooth $\ell_0$-norm constraint introduces significant theoretical and computational challenges. The proposed models belong to the class of combinatorial optimization problems, for which finding global minimizers is generally intractable. To address this, we adopt a penalty decomposition framework whose subproblems are either convex or admit closed-form solutions, depending on the choice of loss function. For the hinge loss, the main subproblem reduces to a linearly constrained quadratic program, which can be efficiently solved via duality theory. In the case of quadratic loss, the main subproblem simplifies to solving a system of linear equations, for which a closed-form solution is obtained. Specifically, Algorithm~\ref{algo: L0-penalty} is a direct application of the general penalty decomposition method proposed in \cite{lu2013sparse}, applied to the context of $\ell_0$-regularized kernel-free quadratic SVMs. This framework circumvents the combinatorial difficulty of direct $\ell_0$-minimization by solving a sequence of tractable subproblems, each enforcing sparsity in a relaxed yet principled manner.

In the subsequent analysis, we demonstrate that Algorithm \ref{algo: L0-penalty} converges to a solution satisfying the Lu-Zhang stationarity conditions--a generalized notion of first-order optimality tailored to nonconvex problems with cardinality constraints. This concept extends the classical KKT framework and is particularly suitable for settings where conventional convex stationarity conditions fail to capture meaningful structure. To formally analyze convergence, we consider a general class of sparse optimization problems with structural constraints, as described in problem~\eqref{pr:Lu-Zhang-problem}. These problems are motivated by applications in signal recovery, image reconstruction, and data compression \cite{mousavi2020survey} and can be formulated as:
\begin{equation}\label{pr:Lu-Zhang-problem} \tag{${P}$}
\min_{{v} \in \mathbb{R}^{\hat{\jmath}}} \,{\varphi}({v}) \qquad \text{subject to} \qquad {\psi}({v}) \le 0, \quad {\chi}({v}) = 0, \quad \|{v}\|_0 \le k,
\end{equation}
where ${\varphi}$, ${\psi}$, and ${\chi}$ are continuously differentiable functions, though not necessarily convex, and $k < \hat{\jmath}$ denotes the desired sparsity level.

A point ${v}$ satisfying ${\psi}({v}) \le 0$ and ${\chi}({v}) = 0$ is called a \emph{Lu-Zhang stationary point} for problem~\eqref{pr:Lu-Zhang-problem} if there exists an index set $\mathcal{L} \subseteq [\hat{\jmath}]$ with $|\mathcal{L}| = k$ such that ${v}_j = 0$ for all $j \in \mathcal{L}^c := [\hat{\jmath}] \setminus \mathcal{L}$, and there exist multipliers ${\lambda} \in \mathbb{R}^{\tilde{q}}$ and ${\mu} \in \mathbb{R}^{\check{r}}$ satisfying:
\begin{equation}\label{e.Lu-Zhang_stationary}
\begin{cases}
\nabla{\varphi}({v}) - {\lambda}^{\top} \nabla {\psi}({v}) - {\mu}^{\top} \nabla {\chi}({v}) - {\omega} = 0, \\
{\lambda}_i \ge 0, \quad {\lambda}_i \cdot {\psi}_i({v}) = 0, \quad \forall i, \quad \text{and} \quad {\omega}_j = 0, \quad \forall j \in \mathcal{L}^c.
\end{cases}
\end{equation}
While alternative stationarity concepts exist (e.g., basic feasibility \cite{beck2016minimization}), Lu-Zhang stationarity provides a suitable theoretical foundation for convergence analysis in nonconvex sparse settings. It is known that basic feasible points are also Lu-Zhang stationary points, though the converse does not always hold \cite{lapucci2021convergent}.

To underscore the importance of Lu-Zhang stationarity, recall that Theorem 2.1 in \cite{lu2013sparse} establishes that any local minimizer of problem~\eqref{pr:Lu-Zhang-problem} satisfies the Lu-Zhang stationarity condition under Robinson’s constraint qualification.
Furthermore, when ${\varphi}$ and ${\psi}$ are convex and ${\chi}$ is affine, any Lu-Zhang stationary point with support of size exactly $k$ is a local minimizer. In our case, both proposed models—\ref{L0-QSVM'} and \ref{LS-L0-QSVM'}—satisfy these assumptions: the objective functions are convex (since $G \succeq 0$), and the structural constraints are linear. Under such linearity, Robinson’s condition is automatically satisfied. Therefore, it suffices to show that Algorithm~\ref{algo: L0-penalty} converges to a Lu-Zhang stationary point, which implies that the obtained solution satisfies the necessary conditions for local optimality. Moreover, when such a solution has full support (i.e., $\|x\|_0 = k$), it is in fact a local minimizer of the corresponding model.

Next, observe that Algorithm~\ref{algo: L0-penalty} is a direct application of the general penalty decomposition framework developed in \cite{lu2013sparse}, and thus inherits its convergence guarantees without further modification. In particular, Theorem 4.3 in \cite{lu2013sparse} guarantees convergence to a Lu-Zhang stationary point, provided that Robinson’s constraint qualification holds—an assumption automatically satisfied in our models due to the linearity of their constraint sets $\mathbb{K}$. Furthermore, since the objective functions in both formulations are convex, any Lu-Zhang stationary point with full support of size $k$ is a local minimizer. Hence, Algorithm~\ref{algo: L0-penalty} produces feasible $k$-sparse solutions where the gradient of the Lagrangian vanishes on the support, certifying convergence to a Lu-Zhang stationary point of problem~\eqref{generic-L0-quadratic}.
More specifically, in solving the hinge loss model \ref{L0-QSVM'}, Algorithm~\ref{algo: L0-penalty} obtains a point $(z, c, \xi) \in \mathbb{R}^{\frac{n(n+1)}{2}+n} \times \mathbb{R} \times \mathbb{R}^m$ such that $z$ is supported on a set $\mathcal{L} \subseteq \left[\frac{n(n+1)}{2}+n\right]$ with $|\mathcal{L}| = k$, and the following conditions hold:
\[
\left\{
\begin{array}{ll}
G z - \sum_{i=1}^m \lambda_i y_i r_i - {\omega} = 0, &  \text{where } {\omega}_j = 0 \text{ for } j \in \mathcal{L} \\[4pt]
\sum_{i=1}^m \lambda_i y_i = 0, & \\[4pt]
{\mathcal{C}} - \lambda_i - \bar{\lambda}_i = 0, & \forall i \in [m] \\[4pt]
\lambda_i \ge 0, \quad \lambda_i \cdot \left( y_i(z^{\top} r_i + c) + \xi_i - 1 \right) = 0, & \forall i \in [m] \\[4pt]
\bar{\lambda}_i \ge 0, \quad \bar{\lambda}_i \cdot \xi_i = 0, & \forall i \in [m] \\[4pt]
y_i(z^{\top} r_i + c) \ge 1 - \xi_i, \quad \xi_i \ge 0, & \forall i \in [m]
\end{array}
\right.
\]

Furthermore, when solving the quadratic loss model \ref{LS-L0-QSVM'}, Algorithm~\ref{algo: L0-penalty} returns a point $(z, c, \xi)$ satisfying $z_j = 0$ for all $j \in \mathcal{L}^c := \left[\frac{n(n+1)}{2}+n\right] \setminus \mathcal{L}$ with $|\mathcal{L}|=k$, and the following Lu-Zhang stationarity conditions hold:
\[
\left\{
\begin{array}{ll}
G z - \sum_{i=1}^m \mu_i y_i r_i - {\omega} = 0, & \text{where } {\omega}_j = 0 \text{ for } j \in \mathcal{L} \\[4pt]
\sum_{i=1}^m \mu_i y_i = 0, & \\[4pt]
2{\mathcal{C}} \xi_i + \mu_i = 0, & \forall i \in [m] \\[4pt]
y_i(z^{\top} r_i + c) = 1 - \xi_i, & \forall i \in [m].
\end{array}
\right.
\]

\section{Numerical Experiments} \label{sec: numerical_experiments}
In this section, we conduct numerical experiments to evaluate the effectiveness of the proposed \ref{L0-QSVM} and \ref{LS-L0-QSVM} models for classification. We first describe the experimental settings and the benchmark models used for comparison. In Section \ref{section: benchmark datasets}, we present results on several public benchmark datasets. Subsequently, in Section \ref{section: credit scoring}, we apply the relatively more efficient \ref{LS-L0-QSVM} model to credit scoring using real-world credit datasets.

The implementation of the proposed models \ref{L0-QSVM} and \ref{LS-L0-QSVM} follows Algorithm \ref{algo: L0-penalty}, guided by the instructions in Corollaries \ref{cor: 1} and \ref{cor: 2} for Step 6. For comparison, we implement several widely used SVM-based models, including \ref{QSVM}, linear SVM (LSVM), least squares SVM (LS-SVM), radial basis function kernel SVM (SVM-rbf), and quadratic kernel SVM (SVM-Q2). In addition, we include several sparse SVM models, including the linear $\ell_1$-regularized SVM ($\ell_1$-SVM), the quadratic $\ell_1$-regularized SVM (\ref{L1-QSVM}), and the least squares $\ell_0$-regularized SVM (LS-$\ell_0$-SVM) \cite{tang2024sparse}.


\begin{table}[h]
    \centering
    \begin{tabular}{ccc}
    \toprule 
    Model & Algorithm/Package/Solver & Parameters \\
    \midrule 
    LSVM & Scikit-learn  & $\mathcal C$ \\
    LS-SVM & Numpy  & $\mathcal C$ \\
    $\ell_1$-SVM & COPT & $\mathcal C$ \\
    LS-$\ell_0$-SVM & ADMM & $\mathcal C, \gamma, \lambda$ \\
    SVM-rbf & Scikit-learn & $\mathcal C, \gamma$ \\
    SVM-Q2 & Scikit-learn & $\mathcal C, \gamma$ \\
    \ref{QSVM} & COPT & $\mathcal C$ \\
    \ref{L1-QSVM} & COPT &  $\mathcal C_1, \mathcal C_2$ \\
    $\ell_0$-QSVM & Algorithm \ref{algo: L0-penalty}, COPT &  $\mathcal C, k$ \\
    LS-$\ell_0$-QSVM & Algorithm \ref{algo: L0-penalty}, \eqref{sol-Pz-quad}, COPT &  $\mathcal C, k$ \\
    \bottomrule 
    \end{tabular}
    \caption{Information of Implemented Models}
    \label{tab: model details}
\end{table}


All experiments are implemented in Python 3.9.12 and performed on a MacBook Pro equipped with an Apple M3 Pro Chip and 18GB of Memory. The Cardinal Optimizer (COPT) solver in version 8.0.2 is employed for solving the optimization problems in the implementation. Python package  \textit{scikit-learn} \cite{pedregosa2011scikit} is also adopted to implement linear and kernel-based SVM models.

\subsection{Benchmark Datasets}
\label{section: benchmark datasets}

All models are implemented on the datasets listed in Table \ref{tab:results benchmark}, which are widely used benchmarks for assessing newly developed methods in the literature. All datasets are binary except the \textit{iris} dataset, which contains three classes. For the multi-class setting, we adopt a one-vs-rest scheme and assign the predicted label by majority voting. The grid search with a five-fold cross validation is performed to tune the parameters of all implemented models. The mean accuracy and standard deviation is recorded for each model on each dataset in Table \ref{tab:results benchmark}.


\begin{table}[h]
\centering
\caption{Results on Public Benchmark Datasets}
\label{tab:results benchmark}
\resizebox{\columnwidth}{!}{%
\begin{tabular}{lccccccccccccccc}
        \toprule
\multicolumn{2}{l}{\multirow{3}{*}{ Model }} &  \multicolumn{2}{c}{ abalone } & \multicolumn{2}{c}{ CTG } & \multicolumn{2}{c}{ Ecoli } & \multicolumn{2}{c}{ glass } & \multicolumn{2}{c}{ haberman } & \multicolumn{2}{c}{ Immunotherapy } & \multicolumn{2}{c}{ iris }    \\ 
\multicolumn{2}{c}{}                   &   \multicolumn{2}{c}{$( 2854, 8 )$ }  & \multicolumn{2}{c}{$( 2126, 22)$ }  & \multicolumn{2}{c}{$( 336, 7)$ }  & \multicolumn{2}{c}{$( 214, 9)$ }  & \multicolumn{2}{c}{$( 306, 3)$ }  & \multicolumn{2}{c}{$( 90, 7)$ }  & \multicolumn{2}{c}{$( 150, 4)$ }   \\ \cline{3-16}
\multicolumn{2}{c}{}                   &   acc & f1 & acc & f1 & acc & f1 & acc & f1 & acc & f1 & acc & f1 & acc & f1  \\ 
\midrule
\multirow{2}{*}{ LSVM } &  mean & 87.21 & 87.18 & 96.90 & 98.02 & 97.02 & 96.48 & 92.97 & 84.25 & 73.57 & 17.17 & 81.11 & 89.06 & 96.67 & 96.80  \\ 
  &  std & 1.31 & 0.69 & 0.56 & 0.37 & 1.83 & 2.31 & 2.96 & 4.86 & 9.12 & 18.32 & 7.45 & 4.79 & 3.33 & 3.10  \\ 
\multirow{2}{*}{ LS-SVM } &  mean & 86.51 & 86.48 & 95.16 & 96.98 & 97.03 & 96.55 & 92.04 & 78.17 & 73.57 & 17.17 & 80.00 & 88.47 & 83.33 & 83.59  \\ 
  &  std & 1.43 & 0.86 & 0.79 & 0.48 & 1.48 & 1.65 & 3.20 & 16.21 & 9.12 & 18.32 & 8.43 & 5.44 & 8.50 & 6.31  \\ 
\multirow{2}{*}{ SVM-rbf } &  mean & \textbf{89.70} & \textbf{89.95} & \textbf{98.50} & \textbf{99.03} & 97.02 & 96.48 & \textbf{94.39} & \textbf{86.75} & 71.28 & 27.37 & 78.89 & 87.92 & 96.67 & 97.02  \\ 
  &  std & 1.14 & 0.75 & 0.35 & 0.24 & 1.06 & 1.40 & 2.11 & 8.36 & 9.68 & 19.29 & 9.94 & 6.25 & 4.71 & 4.11  \\ 
\multirow{2}{*}{ SVM-Q2 } &  mean & 89.28 & 89.43 & 98.31 & 98.91 & 97.03 & 96.60 & 93.92 & 85.56 & 72.58 & 18.06 & 80.00 & 88.19 & 96.67 & 97.19  \\ 
  &  std & 1.08 & 0.70 & 0.58 & 0.38 & 1.03 & 0.87 & 3.90 & 12.17 & 6.92 & 13.00 & 10.83 & 6.81 & 5.77 & 4.64  \\ 
\multirow{2}{*}{ $\ell_1$-SVM } &  mean & 87.70 & 87.73 & 97.18 & 98.19 & 96.43 & 95.80 & 92.98 & 84.32 & 71.28 & 3.33 & 80.00 & 88.27 & 96.67 & 96.80  \\ 
  &  std & 1.10 & 0.53 & 0.55 & 0.35 & 1.32 & 1.58 & 2.38 & 3.71 & 8.17 & 7.45 & 11.52 & 7.23 & 3.33 & 3.10  \\ 
\multirow{2}{*}{ LS-$\ell_0$-SVM } &  mean & 86.69 & 86.69 & 95.39 & 97.12 & 97.03 & 96.55 & 92.04 & 78.17 & 73.57 & 17.17 & 80.00 & 88.47 & 84.00 & 83.95  \\ 
  &  std & 1.60 & 1.22 & 0.56 & 0.34 & 1.48 & 1.65 & 3.20 & 16.21 & 9.12 & 18.32 & 8.43 & 5.44 & 5.48 & 4.09  \\ 
\multirow{2}{*}{ \ref{QSVM} } &  mean & 88.89 & 89.11 & 98.21 & 98.85 & 97.03 & 96.60 & 92.98 & 83.68 & 72.58 & 18.06 & 81.11 & 88.79 & 97.33 & 97.43  \\ 
  &  std & 1.50 & 1.01 & 0.21 & 0.15 & 1.03 & 0.87 & 4.68 & 14.01 & 6.92 & 13.00 & 11.52 & 7.17 & 2.79 & 2.77  \\ 
\multirow{2}{*}{ \ref{L1-QSVM}} &  mean & 88.79 & 88.99 & 98.21 & 98.85 & 97.03 & 96.60 & 94.39 & 86.49 & 73.56 & 19.37 & 81.11 & 88.83 & 97.33 & 97.32  \\ 
  &  std & 1.47 & 0.98 & 0.27 & 0.18 & 1.03 & 0.87 & 3.54 & 12.14 & 6.85 & 13.60 & 11.52 & 7.27 & 1.49 & 1.55  \\ 
\multirow{2}{*}{ \ref{L0-QSVM}} &  mean & 89.21 & 89.32 & 98.40 & 98.97 & 97.02 & 96.48 & 93.44 & 86.48 & 73.24 & 19.11 & 81.11 & 88.85 & \textbf{97.33} & \textbf{97.56}  \\ 
  &  std & 1.44 & 1.03 & 0.35 & 0.24 & 1.06 & 1.40 & 3.86 & 4.17 & 6.98 & 13.18 & 10.83 & 6.47 & 3.65 & 3.37  \\ 
\multirow{2}{*}{ \ref{LS-L0-QSVM}} &  mean & 88.44 & 88.54 & 96.24 & 97.62 & \textbf{97.33} & \textbf{96.98} & 93.43 & 85.27 & \textbf{74.22} & \textbf{32.10} & \textbf{83.33} & \textbf{89.56} & 96.67 & 97.00  \\ 
  &  std & 1.65 & 1.18 & 0.92 & 0.58 & 1.62 & 1.70 & 3.55 & 8.54 & 8.10 & 11.55 & 14.16 & 8.94 & 5.77 & 4.68  \\ 
    \bottomrule
\end{tabular}%
}
\end{table}


From the results, we can observe that the proposed \ref{L0-QSVM} and the \ref{LS-L0-QSVM} provide the highest in both accuracy and f1-score on the \textit{Ecoli}, \textit{haberman}, \textit{Immunotherapy} and \textit{Iris} datasets. Specifically, the \ref{LS-L0-QSVM} model outperforms \ref{L0-QSVM} on most of these cases. Overall, the proposed models are consistently competitive with commonly used linear and kernel-based SVM benchmark models.


\begin{figure}[h]
\centering
\begin{subfigure}{.3\textwidth}
  \centering
  \includegraphics[width=\linewidth]{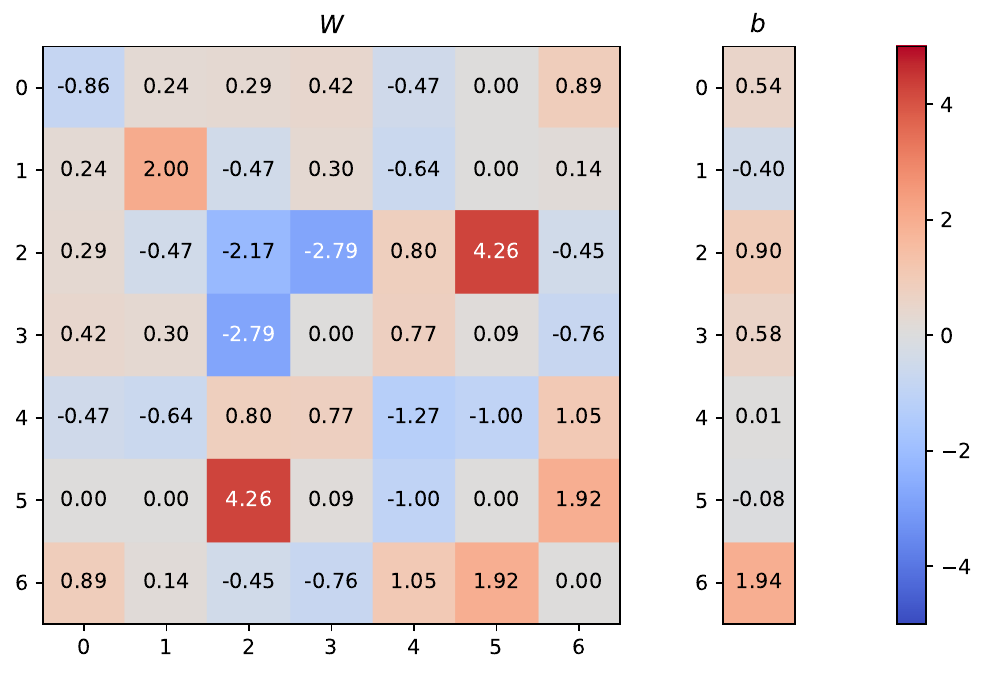}
  \caption{\ref{L1-QSVM} \\ ($\mathcal C_1 = 256, \mathcal C_2 = 100$)}
  \label{fig: L1Q}
\end{subfigure}%
\begin{subfigure}{.3\textwidth}
  \centering
  \includegraphics[width=\linewidth]{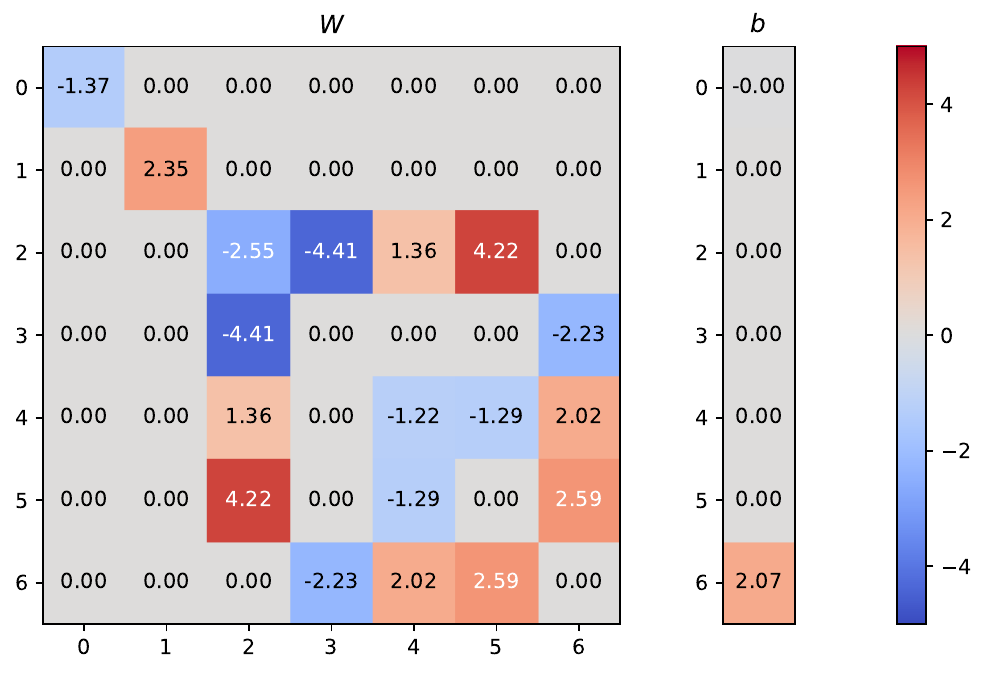}
    \caption{\ref{L0-QSVM} \\ ($\mathcal C = 1024, k = 12$)}
  \label{fig: L0Q}
\end{subfigure}%
\begin{subfigure}{.3\textwidth}
  \centering
  \includegraphics[width=\linewidth]{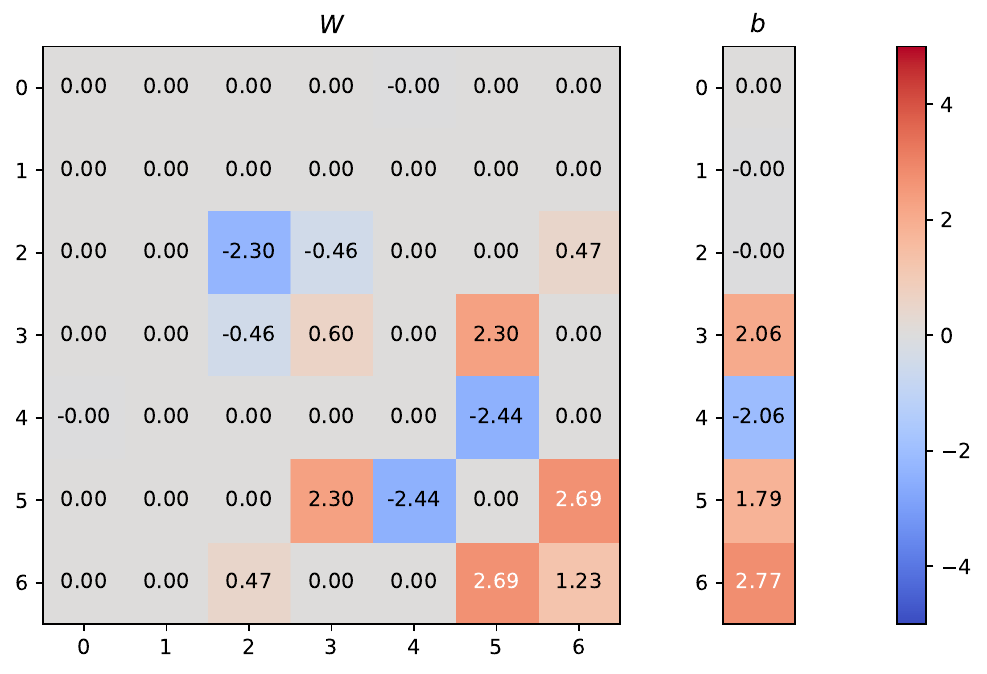}
  \caption{\ref{LS-L0-QSVM} \\ ($\mathcal C = 1024, k = 12$)}
  \label{fig: LSL0Q}
\end{subfigure}
\caption{The sparsity of optimal coefficients $W^*$ and $b^*$ by three sparse quadratic models on the \textit{Immunotherapy} dataset.}

\label{fig: sparsity W b immunotherapy}
\end{figure}


The proposed \ref{L0-QSVM} and \ref{LS-L0-QSVM} models produce sparse solutions, as expected from their formulation. To illustrate this property, we visualize the optimal coefficients $(W^, b^)$ of the two proposed models and the \ref{L1-QSVM} model \cite{mousavi2022quadratic} on the \textit{Immunotherapy} dataset in Figure \ref{fig: sparsity W b immunotherapy}. It is evident that both \ref{L0-QSVM} and \ref{LS-L0-QSVM} yield well-controlled sparsity patterns. Although \ref{LS-L0-QSVM} produces a different sparsity structure in $(W^*, b^*)$ compared with \ref{L0-QSVM}, this difference stems from their distinct classification mechanisms.

In contrast, while increasing the penalty parameter $\mathcal C_2$ in \ref{L1-QSVM} can encourage sparsity of solutions, it does not allow precise control over the number of nonzero coefficients. As a result, it is difficult to obtain solutions with a predetermined sparsity level using \ref{L1-QSVM}. By comparison, the proposed \ref{L0-QSVM} and \ref{LS-L0-QSVM} models enable direct control of sparsity, allowing solutions with an explicitly specified number of nonzero coefficients.


\begin{figure}[h]
\centering
\begin{subfigure}{.4\textwidth}
  \centering
  \includegraphics[width=\linewidth]{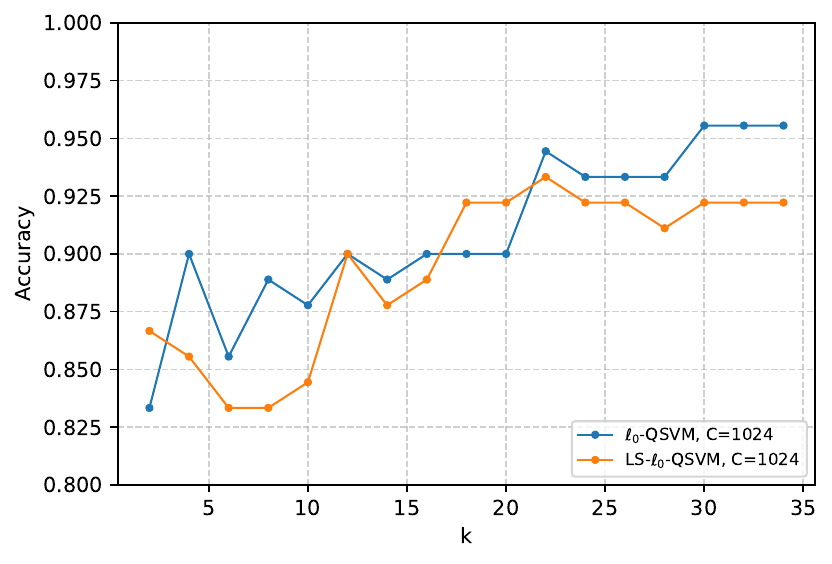}
  \caption{\textit{Immunotherapy}: Accuracy vs. $k$ \\ ($\mathcal C = 1024$)}
  \label{fig: Immunotherapy Acc vs k}
\end{subfigure}%
\begin{subfigure}{.4\textwidth}
  \centering
  \includegraphics[width=\linewidth]{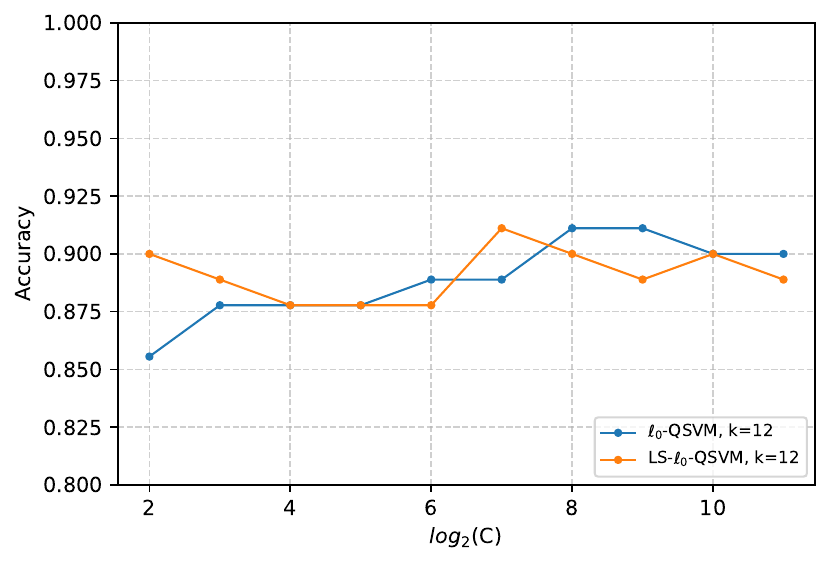}
  \caption{\textit{Immunotherapy}: Accuracy vs. $C$ \\ ($k = 12$)}
  \label{fig: Immunotherapy Acc vs C}
\end{subfigure}

\begin{subfigure}{.4\textwidth}
  \centering
  \includegraphics[width=\linewidth]{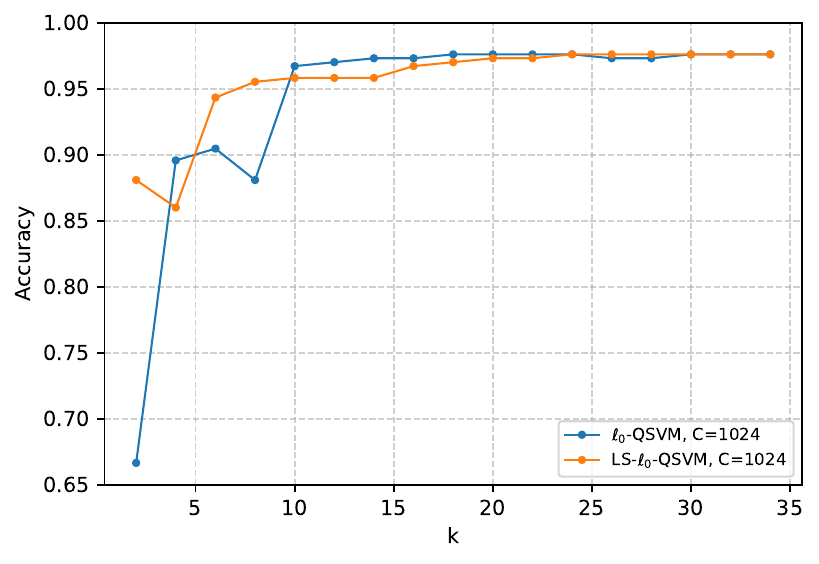}
  \caption{\textit{Ecoli}: Accuracy vs. $k$ \\ ($\mathcal C = 1024$)}
  \label{fig: Ecoli Acc vs k}
\end{subfigure}%
\begin{subfigure}{.4\textwidth}
  \centering
  \includegraphics[width=\linewidth]{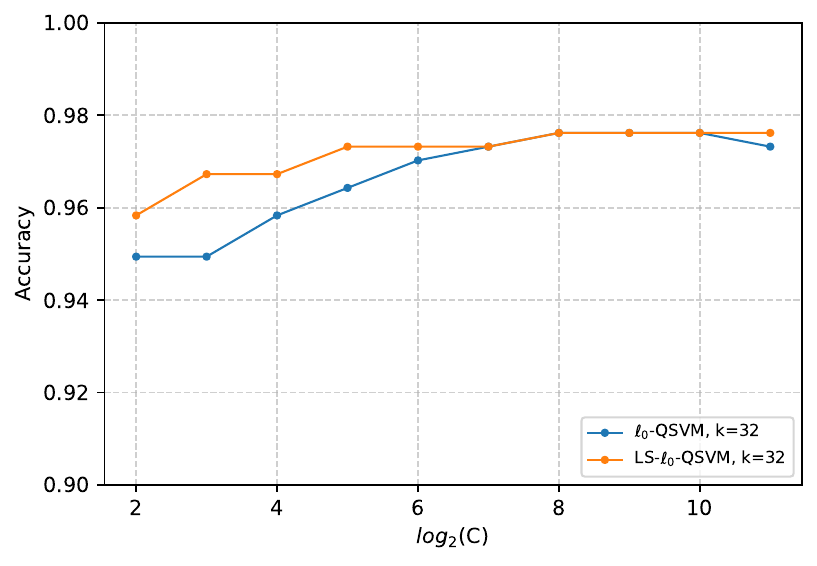}
  \caption{\textit{Ecoli}: Accuracy vs. $C$ \\ ($k = 32$)}
  \label{fig: Ecoli Acc vs C}
\end{subfigure}

\caption{The trend of accuracy against parameters $C$ and $k$ by the proposed $\ell_0$ regularized QSVM models on the \textit{Immunotherapy} and \textit{Ecoli} datasets.}

\label{fig: trend C, k, benchmark}
\end{figure}


Next, we investigate the effects of the parameters $\mathcal C$ and $k$ on classification performance. Figures \ref{fig: Ecoli Acc vs k} and \ref{fig: Immunotherapy Acc vs k} present the accuracy curves of \ref{L0-QSVM} (blue) and \ref{LS-L0-QSVM} (yellow) on the \textit{Ecoli} and \textit{Immunotherapy} datasets, respectively, by fixing $\mathcal C$ and varying $k$. We observe that when $k$ increases from a small value, the classification accuracy improves steadily. However, after $k$ exceeds a certain threshold, the performance gain becomes marginal, particularly for the \ref{LS-L0-QSVM} model. This behavior indicates that once a sufficient number of informative coefficients are selected, adding additional nonzero terms contributes little to further improvement. On both datasets, especially \textit{Ecoli}, accuracy increases rapidly for small $k$, then the curve becomes flat. This indicates that only a relatively small subset of interacted and linear terms is necessary to achieve an optimal performance. This strongly supports the effectiveness of sparsity control.

Similarly, Figures \ref{fig: Ecoli Acc vs C} and \ref{fig: Immunotherapy Acc vs C} illustrate the accuracy curves obtained by fixing $k$ and varying $\mathcal C$. The results show that the performance of both models remains relatively stable across a wide range of $\mathcal C$ values, provided that $k$ is appropriately chosen. This suggests that the proposed models are not highly sensitive to the penalty parameter $\mathcal C$, and that sparsity control through $k$ plays a more dominant role in determining predictive performance.

\subsection{Application: Credit Scoring}
\label{section: credit scoring}

Credit scoring plays a crucial role in financial institutions by supporting risk assessment, underwriting decisions, and portfolio management. Accurate credit scoring helps lenders reduce default risk, improve pricing strategies, and enhance operational efficiency and financial stability. Traditional credit scoring approaches, such as expert-driven scorecards and statistical models \cite{mays1995handbook, splett1994joint}, are largely based on handcrafted features and domain expertise, which may limit their flexibility when facing complicated behaviors of borrowers.

Credit scoring can be naturally formulated as a classification problem, where the objective is to assign each applicant to a risk category based on observed attributes. In recent years, machine learning methods have been increasingly adopted for credit scoring, including decision trees \cite{dumitrescu2022machine}, logistic regression \cite{bolton2009logistic}, and neural networks \cite{west2000neural}, due to their ability to capture nonlinear relationships and feature interactions. Among these methods, SVM have demonstrated competitive performance in credit scoring applications \cite{huang2007credit}. In this section, we use five publicly available credit scoring datasets to evaluate the effectiveness of the proposed LS-$\ell_0$-QSVM model for credit scoring. Five-fold cross validation is also applied.


\begin{table}[h]
\centering
\caption{Credit Data Information}
\label{tab: credit data information}
\resizebox{\columnwidth}{!}{%
\begin{tabular}{lcccl}
    \toprule
    Dataset & Source & Type & \makecell[c]{Dimension \\ $(m, n)$} & \makecell[c]{Features} \\
    \midrule
        CCC & private credit union & corporation loan & $(106, 6)$ & \makecell[l]{1.\  Number of collaborations or total deliveries within the past three years \\
2.\ Number of integrity violations \\
3.\ Number of delivery-related violations \\
4.\ Number of quality-related violations \\
5.\ Credit China – number of positive records \\
6.\ Credit China – number of blacklist penalties} \\
    \hline
    credit small & Kaggle & personal credit & $(164, 7)$ & \makecell[l]{1.\ Age \\
2.\ Gender \\
3.\ Income \\
4.\ Education \\
5.\ Marital Status \\
6.\ Number of Children \\
7.\ Home Ownership} \\
    \hline
        GCD & UCI & personal credit & $(1000, 20)$ & \makecell[l]{1.\ Status of existing checking account \\
2.\ Duration in month \\
3.\ Credit hisotry \\
4.\ Loan purpose \\
5.\ Credit amount \\
6.\ Status of savings account/bonds \\
7.\ Years of current employment \\
8.\ Installment rate in percentage of disposable income \\
9.\ Gender and marriage status \\
10.\ Other debtors / guarantors \\
11.\ Length of present residence \\
12.\ Property status \\
13.\ Age \\
14.\ Other installment plans \\
15.\ Housing \\
16.\ Number of existing credits at this bank \\
17.\ Job status \\
18.\ Number of people being liable to provide maintenance for \\
19.\ Phone indicator \\
20.\ Foreign worker indicator
} \\
    \hline
        JAP & UCI & personal credit & $(653, 15)$ & Processed numeric variables \\
    \hline
    AUS & UCI & personal credit & $(690, 14)$ & Processed numeric variables \\
        \bottomrule
\end{tabular}
}
\end{table}


We consider five credit datasets from both public repositories and practical industry sources, including the personal and corporate credit datasets with varying sample sizes and features. The basic information of the datasets, including their source and dimension, is listed in Table \ref{tab: credit data information}.  Among them, the \textit{German Credit Dataset} (\textit{GCD}), \textit{Japanese Credit} (\textit{JAP}), and \textit{Australian Credit} (\textit{AUS}) datasets are obtained from the UCI Machine Learning Repository. The \textit{GCD} contains 1000 loan applicants described by 20 attributes. These features include financial indicators, employment information, demographic characteristics, and other credit-related records. The response variable is binary, indicating whether an applicant is classified as a good credit risk or a bad credit risk. Similarly, the \textit{AUS} dataset consists of 690 samples with 14 attributes. It is a famous binary benchmark dataset with Australian credit card applications. The \textit{JAP} dataset contains 653 instances with 15 attributes. For both the \textit{GCD} and \textit{JAP} datasets, the original attribute names and categorical values were replaced with symbolic representations to protect data confidentiality.

In addition to the UCI datasets, we also include two smaller datasets. The \textit{CCC} dataset is obtained from a private credit union and focuses on corporate loan evaluation. It contains 106 samples with 6 interpretable financial and compliance-related features, such as the number of collaborations, integrity violations, and blacklist penalties. The response label indicates the credit status of the corporation. The \textit{credit small} dataset is collected from Kaggle and represents personal credit assessment with 164 samples and 7 demographic and socioeconomic features, including age, income, education, marital status, and home ownership. The response variable is binary and reflects the credit risk level of individuals.

In this application
The proposed LS-$\ell_0$-QSVM model is implemented on the datasets in Table \ref{tab: credit data information} along with benchmark models. The mean classification accuracies and the f1-scores are recorded in Table \ref{tab: results on credit data}.


\begin{table}[h]
\centering
\caption{Results on Credit Data}
\label{tab: results on credit data}
\resizebox{0.8\columnwidth}{!}{%
\begin{tabular}{lccccccccccc}
        \toprule
\multicolumn{2}{l}{\multirow{2}{*}{Model}} &   \multicolumn{2}{c}{  AUS  }  &  \multicolumn{2}{c}{  JAP  }  &  \multicolumn{2}{c}{  GCD  }  &  \multicolumn{2}{c}{  CCC  }  &  \multicolumn{2}{c}{  credit small  }     \\ \cline{3-12}
\multicolumn{2}{c}{}                     &    acc  &  f1  &  acc  &  f1  &  acc  &  f1  &  acc  &  f1  &  acc  &  f1   \\ 
    \midrule
\multirow{2}{*}{  LSVM  } &     mean &  87.10   &  87.83   &  86.83   &  86.56   &  76.90   &  55.79   &  89.70   &  85.23   &  98.18   &  98.13    \\ 
    &   std &  1.88   &  2.05   &  1.60   &  1.42   &  1.85   &  2.47   &  9.57   &  13.71   &  2.71   &  2.82    \\ 
\multirow{2}{*}{  LS-SVM  } &     mean &  86.38   &  86.82   &  86.67   &  86.46   &  76.80   &  54.81   &  87.75   &  84.24   &  90.87   &  76.97    \\ 
    &   std &  2.37   &  2.92   &  1.71   &  1.66   &  2.11   &  2.99   &  5.38   &  8.16   &  8.55   &  19.36    \\ 
\multirow{2}{*}{  SVM-rbf  } &     mean &  86.67   &  87.27   &  86.21   &  85.93   &  76.10   &  54.53   &  90.56   &  86.49   &  98.18   &  98.13    \\ 
    &   std &  1.75   &  2.25   &  1.36   &  1.71   &  3.38   &  6.71   &  7.53   &  11.95   &  2.71   &  2.82    \\ 
\multirow{2}{*}{  SVM-Q2  } &     mean &  86.23   &  87.39   &  86.37   &  86.12   &  75.70   &  56.56   &  \textbf{92.47}   &  \textbf{89.43}   &  98.18   &  98.13    \\ 
    &   std &  3.28   &  3.46   &  1.40   &  1.59   &  3.29   &  5.86   &  7.94   &  11.85   &  2.71   &  2.82    \\ 
\multirow{2}{*}{  $\ell_1$-SVM  } &     mean &  85.51   &  85.93   &  86.67   &  86.45   &  77.20   &  55.85   &  91.56   &  88.27   &  98.18   &  98.13    \\ 
    &   std &  1.85   &  2.07   &  1.42   &  1.77   &  1.25   &  2.95   &  8.41   &  12.06   &  2.71   &  2.82    \\ 
\multirow{2}{*}{  LS-$\ell_0$-SVM  } &     mean &  86.23   &  86.69   &  86.67   &  86.46   &  77.20   &  55.75   &  87.75   &  84.24   &  90.87   &  76.97    \\ 
    &   std &  2.51   &  2.83   &  1.71   &  1.66   &  1.72   &  2.67   &  5.38   &  8.16   &  8.55   &  19.36    \\ 
\multirow{2}{*}{  \ref{QSVM}  } &     mean &  85.94   &  87.09   &  86.52   &  86.31   &  76.90   &  57.31   &  91.56   &  88.27   &  97.58   &  97.66    \\ 
    &   std &  2.64   &  2.50   &  1.53   &  1.59   &  2.10   &  5.07   &  8.41   &  12.06   &  2.54   &  2.62    \\ 
\multirow{2}{*}{  \ref{L1-QSVM}  } &     mean &  86.38   &  87.53   &  86.52   &  86.31   &  77.40   &  57.42   &  91.56   &  88.27   &  98.18   &  98.13    \\ 
    &   std &  2.87   &  2.95   &  1.53   &  1.59   &  0.55   &  2.25   &  8.41   &  12.06   &  2.71   &  2.82    \\ 
\multirow{2}{*}{  \ref{LS-L0-QSVM} } &     mean &  \textbf{87.25}   &  \textbf{88.28}   &  \textbf{87.14}   &  \textbf{86.59}   &  \textbf{77.50}   &  \textbf{57.48}   &  91.56   &  88.27   &  \textbf{99.39}   &  \textbf{99.33}    \\ 
    &   std &  1.67   &  2.06   &  1.63   &  1.85   &  1.73   &  5.50   &  8.41   &  12.06   &  1.36   &  1.49    \\ 
    \bottomrule
\end{tabular}%
}
\end{table}


From Table \ref{tab: results on credit data}, we observe that the proposed \ref{LS-L0-QSVM} model achieves the highest mean accuracy and F1-scores on most of the tested credit datasets. Although it does not outperform SVM-Q2 on the \textit{CCC} dataset, its performance remains highly competitive. These results provide strong empirical evidence that the proposed \ref{LS-L0-QSVM} model is effective across diverse credit evaluation scenarios.

We further examine the impact of the sparsity parameter $k$ in the proposed \ref{LS-L0-QSVM} model on the \textit{GCD}, \textit{AUS}, and \textit{CCC} datasets. Figure \ref{fig: acc vs k, credit data} illustrates the classification accuracy as a function of $k$ while fixing $\mathcal C$.



\begin{figure}
\centering
\begin{subfigure}{.3\textwidth}
  \centering
  \includegraphics[width=\linewidth]{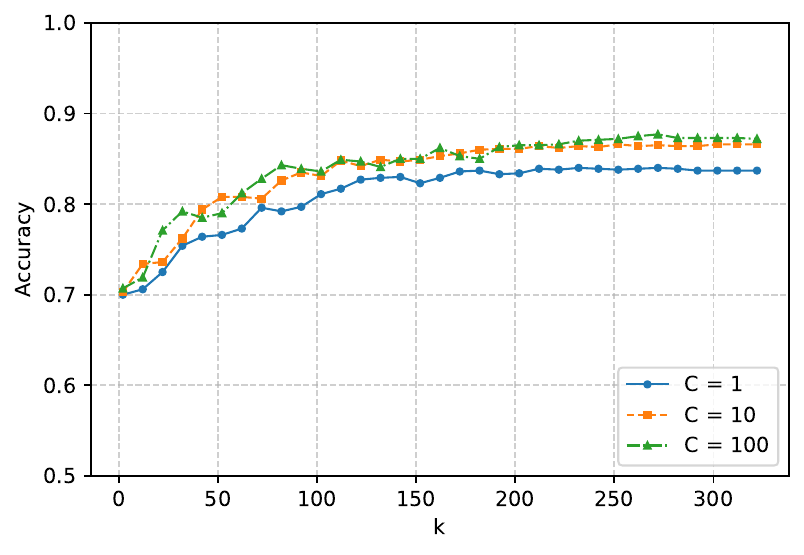}
  \caption{GCD}
  \label{fig: acc vs k, GCD}
\end{subfigure}%
\begin{subfigure}{.3\textwidth}
  \centering
  \includegraphics[width=\linewidth]{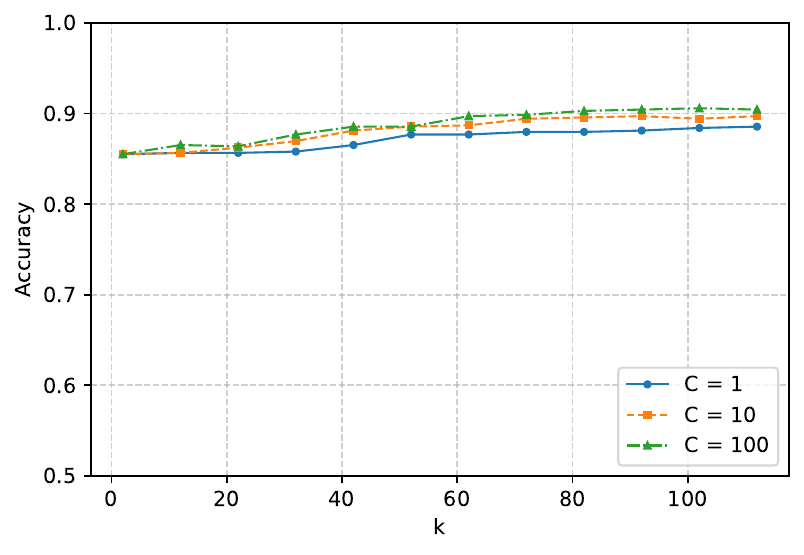}
    \caption{AUS}
  \label{fig: acc vs k, AUS}
\end{subfigure}
\begin{subfigure}{.3\textwidth}
  \centering
  \includegraphics[width=\linewidth]{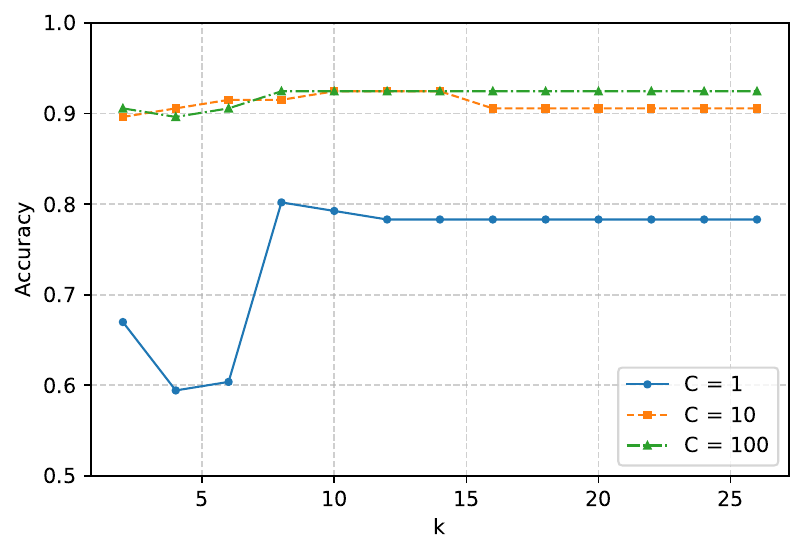}
  \caption{CCC}
  \label{fig: acc vs k, CCC}
\end{subfigure}
\caption{The trend of classification accuracy from \ref{LS-L0-QSVM} changes as $k$ increases. Parameter $\mathcal C$ is fixed to be $1, 10 $ or $100$.}
\label{fig: acc vs k, credit data}
\end{figure}


For the \textit{CCC} and \textit{AUS} datasets, the classification accuracy remains relatively stable across a wide range of $k$ values once an appropriate $\mathcal C$ is selected, indicating that performance is not highly sensitive to the exact sparsity level. In contrast, on the \textit{GCD} dataset, the accuracy initially improves as $k$ increases and then gradually reaches a plateau. This behavior suggests that adopting more features in the model benefits performance up to a certain point, beyond which further increases in model complexity provide negligible improvement in classification accuracy.

\begin{figure}[h]
\centering
  \centering
  \includegraphics[width=0.6\linewidth]{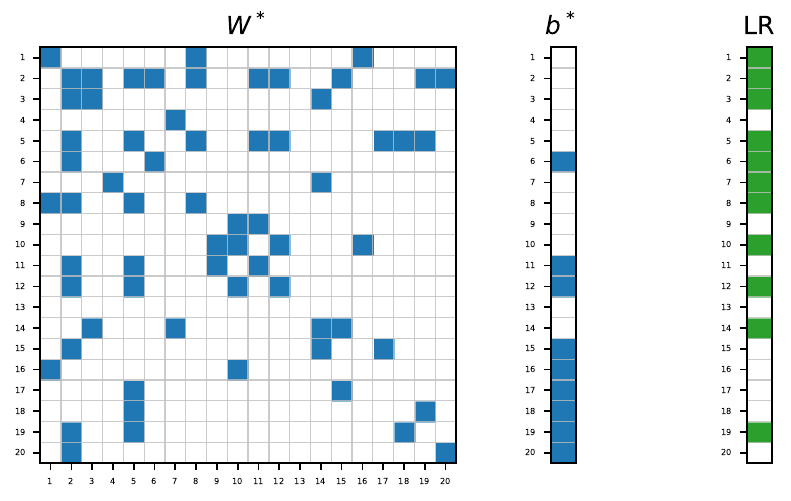}
  \caption{$W^*$ and $b^*$ are sparse optimal solutions provided by \ref{LS-L0-QSVM} with non-zero entries colored in blue. Significant coefficients by LR that are statistically significant at the 95\% confidence level are highlighted in green.}
  \label{fig: significant logistic regression coeff for GCD credit data}
\end{figure}

We further analyze the sparsity of the proposed \ref{LS-L0-QSVM} model on the \textit{GCD} data. By the cross-validation results in Table \ref{tab: results on credit data}, the optimal parameters of the proposed \ref{LS-L0-QSVM} model are $(\mathcal C, k) = (256, 47)$. The corresponding optimal solution $(W^*, b^*)$ is plot in Figure \ref{fig: significant logistic regression coeff for GCD credit data} along with the result of the logistic regression (LR) model on the \textit{GCD} data. In the plot, the non-zero entries of the solution from \ref{LS-L0-QSVM} are highlighted in blue, and the coefficients that are statistically significant at the 95\% confidence level in the LR model are highlighted in green.


We have the following observations from Figure \ref{fig: significant logistic regression coeff for GCD credit data}. 

\begin{enumerate}
    \item The proposed \ref{LS-L0-QSVM} model tells that the credit risk is not driven by single features alone, but by how the core financial variables combine with applicant profile and repayment burdens. Many of these affects are not shown in optimal $b^*$, such as the \textit{Duration (2)}, the \textit{Credit amount (5)} and the \textit{Installment rate (8)}, they appear in the interactions in the optimal quadratic coefficient $W^*$. These features indicates that the default risk is better characterized by interactions between core financial terms and applicant context. 
    
    \item In contrast, the linear vector $b^*$ retains a smaller set of main effects, including \textit{Savings account/bonds (6)}, \textit{Length of present residence (11)}, \textit{Property status (12)}, \textit{Housing (15)}, \textit{Number of existing credits (16)}, \textit{Job status (17)}, \textit{Dependents (18)}, \textit{Phone indicator (19)}, and \textit{Foreign worker (20)}, suggesting these factors contribute more directly without requiring interaction terms.

    \item Notice that LR identifies a larger number of significant predictors among the financial variables (Features 1 to 8) than among the applicant context variables (Features 9 to 20). Compared with LR, the proposed \ref{LS-L0-QSVM} model shifts much of the explanatory power of the financial features into the interaction matrix $W^*$, indicating that their effects are primarily conditional on other variables rather than purely linear. For the applicant context features, the proposed \ref{LS-L0-QSVM} model captures their influence through both the linear coefficients $b^*$ and the quadratic terms represented by the diagonal elements of $W^*$.
\end{enumerate}

The empirical results demonstrate that the proposed \ref{LS-L0-QSVM} model is well-suited for credit scoring applications, and has practical value for real-world credit risk assessment and decision-making.

\section{Conclusion} \label{sec: conclusion}
This paper introduces $\ell_0$ regularized kernel free quadratic surface support vector machine models for binary classification. By enforcing an explicit sparsity constraint, the proposed models produce sparse solutions that improve interpretability while mitigating the risk of overfitting. Although these models are inherently intractable due to the combinatorial nature of the $\ell_0$ penalty, we develop an efficient penalty decomposition algorithm for their implementation. For the hinge loss formulation, we exploit duality theory to solve the resulting subproblem at low computational cost. For the quadratic loss, we show that the subproblem admits a closed form solution. We provide a rigorous convergence analysis of the algorithm, establishing a theoretical foundation that supports its practical use. Numerical experiments confirm the effectiveness of the proposed models on classification tasks. We systematically examine their sparsity properties and investigate the influence of key parameters. Finally, the proposed model \ref{LS-L0-QSVM} is applied to real world credit scoring datasets, where it demonstrates strong predictive performance and interpretability in practical credit risk assessment applications.

The work presented in this paper opens several avenues for future investigation. First, integrating the proposed models with the twin SVM framework could yield more effective solutions for multiclass and imbalanced classification problems. Second, developing adaptive strategies for parameter selection would enhance computational efficiency, particularly when scaling to large datasets. Finally, the interpretability inherent in sparse quadratic decision surfaces positions these models as promising tools for high stakes domains such as healthcare and transportation, where datasets are often high dimensional and transparency is critical for informed decision making.

\section*{Declarations}

\subsubsection*{Competing Interests}
The authors declare that they have no financial or non-financial interests that are directly or indirectly related to the work submitted for publication.

\subsubsection*{Funding}
The authors declare that they have not received any funding, financial support, or sponsorship from any organization or agency for the preparation of this work.

\subsubsection*{Generative AI and AI-assisted Technologies in the Writing Process}
During the preparation of this work, the authors used generative AI technologies to improve language and readability with extreme caution. After using this tool, the authors reviewed and edited the content as needed and take full responsibility for the content of the publication.

\printbibliography
\medskip

\appendix

\section{Tuning Parameters}
\label{appendix section: parameters}

For each implemented model, the values of optimal parameter
s are reported through the 5-fold cross validation with grid search in the corresponding ranges. The ranges of tuning parameters are provided as follows.


\begin{itemize}
\item For the proposed \ref{L0-QSVM} and \ref{LS-L0-QSVM} models, the penalty parameter is selected from $\mathcal C \in \{4^{-1}, \dots, 4^8\}$. The sparsity parameter $k$ is chosen from a uniformly spaced sequence starting at 2 and increasing in increments of $\delta$, up to the maximum possible number of coefficients $\frac{n(n+1)}{2}+n$, where $\delta$ is typically selected from $\{2, 5\}$ depending on the dimensionality of the dataset.

\item For LSVM, LS-SVM, $\ell_1$-SVM, LS-$\ell_0$-SVM, and \ref{QSVM}, the penalty parameter is selected from $\mathcal C \in \{2^{-2}, \dots, 2^{10}\}$.

\item For the LS-$\ell_0$-SVM model, the parameters are selected from $\mathcal C \in \{2^{-2}, \dots, 2^{10}\}$, $\gamma \in \{10^2, \dots, 10^4\}$, and $\lambda \in \{10^{-1}, \dots, 10^1\}$.

\item For the kernel-based models SVM-rbf and SVM-Q2, the penalty parameter is selected from $\mathcal C \in \{2^{-2}, \dots, 2^{10}\}$, and the kernel parameter $\gamma$ is set to the default value provided in the \textit{scikit-learn} package.

\item For \ref{L1-QSVM}, the parameters are chosen from $\mathcal C_1 \in \{2^{-2}, \dots, 2^{10}\}$ and $\mathcal C_2 \in \{10^{-1}, \dots, 10^2\}$.

\end{itemize}




\section{Data source}
\label{appendix section: Data and codes}

The sources of the datasets utilized in Sections \ref{sec: numerical_experiments} are available in the GitHub repository.

\url{https://github.com/tonygaobasketball/L0SQSSVM}.

\end{document}